# Computer Vision for Construction Progress Monitoring: A Real-Time Object Detection Approach

Jiesheng Yang[1], Andreas Wilde[2], Karsten Menzel[1], Md Zubair Sheikh[1]
Boris Kuznetsov[3]

[1] TU Dresden, Institut für Bauinformatik, Dresden, Germany
`{Jiesheng.Yang, Karsten.Menzel1, md_zubair.sheikh}@tu-dresden.de`
[2] Fraunhofer-Institut für Integrierte Schaltungen IIS, Dresden, Germany
`andreas.wilde@eas.iis.fraunhofer.de`
[3] Implenia Hochbau GmbH, Leipzig, Germany
`boris.kuznetsov@implenia.com`

**Abstract.** Construction progress monitoring (CPM) is essential for effective project management, ensuring on-time and on-budget delivery. Traditional CPM methods often rely on manual inspection and reporting, which are time-consuming and prone to errors. This paper proposes a novel approach for automated CPM using state-of-the-art object detection algorithms. The proposed method leverages e.g. YOLOv8's real-time capabilities and high accuracy to identify and track construction elements within site images and videos. A dataset was created, consisting of various building elements and annotated with relevant objects for training and validation. The performance of the proposed approach was evaluated using standard metrics, such as precision, recall, and F1-score, demonstrating significant improvement over existing methods. The integration of Computer Vision into CPM provides stakeholders with reliable, efficient, and cost-effective means to monitor project progress, facilitating timely decision-making and ultimately contributing to the successful completion of construction projects.

**Keywords:** AI and digital transformation, Digital Twins.

## 1 Introduction

Construction progress monitoring (CPM) is a critical aspect of effective project management, as it ensures timely delivery and budget adherence. As construction projects become increasingly complex, the importance of accurate and efficient progress monitoring cannot be overstated. Traditional CPM methods largely depend on manual inspection and reporting, which involve site visits, visual observation, and documentation of construction progress. These methods are often time-consuming, labor-intensive, and error-prone, leading to potential delays, or reduced overall project performance. Furthermore, manual CPM methods can also be impacted by human subjectivity, resulting in inconsistencies and inaccuracies in the collected data.

Comprehensive and consistent CPM is a major contribution to quality management in the AECO-sector. The quality of work executed has a substantial impact on buildings' performance and is in some cases a pre-requisite for the introduction of novel, innovative business models, such as energy service contracting [1], [2], predicted building automation and control [3] or performance-based maintenance [4]. Therefore, we selected for our research the monitoring of the installation process of windows as an example. Instantaneous monitoring and fast analysis of monitoring results will enable the quality managers appointed by the property owner to contact the main contractor to request the correction of low-quality installations. Numerous sub-contractors are involved in the installation process. Thus, our paper explains AI-support of a highly collaborative work-process on construction sites.

Early work in the area of digital CPM focused on the deployment of mobile, wearable devices on construction sites [5] and the development flexible, dynamic business-process models [6]. In recent years, there has been a growing interest in leveraging technological advancements to automate CPM [7]. Several approaches were explored, such as image-based techniques, laser scanning, and Building Information Modeling (BIM) [8, 9] or semantic web technologies [10], [11]. However, these methods have their own drawbacks. Image-based techniques typically require manual processing of the collected images. Laser scanning and BIM can be expensive and time-consuming to implement and semantic-web technologies cannot be easily integrated in commercially available solutions. As a result, there is a need for a more efficient, accurate, and cost-effective CPM solutions. Deep learning and computer vision offer opportunities for automating CPM through object detection algorithms. These algorithms can identify and track objects within images and videos, providing real-time information on their location, size, and orientation.

This paper introduces a novel approach to automated CPM using the state-of-the-art object detection algorithm, YOLOv8. Leveraging YOLOv8's real-time capabilities and high accuracy, our method identifies and tracks construction elements within site images. YOLOv8, a single-stage object detection algorithm, outperforms other methods in accuracy and speed, and its architecture enables simultaneous detection of multiple objects, making it ideal for complex construction sites.

The primary objective of this study is to explore the potential of YOLOv8 as an effective tool for automating CPM, by evaluating its performance in detecting and tracking construction elements within a custom dataset of construction site images and videos. The dataset was created by collecting images and videos from various construction projects and annotating them with relevant objects, such as construction materials, or equipment. The performance of the proposed approach was evaluated using standard metrics, e.g. precision or recall, and compared to existing methods.

The remainder of the paper is structured as follows: Section 2 provides a review of the background and related work on CPM and object detection algorithms. Section 3 introduces the YOLOv8 object detection algorithm and its relevance to CPM. Section 4 describes the dataset creation and annotation process, while Section 5 presents the implementation of the proposed CPM method. Section 6 discusses the results and potential implications of the study, and Section 7 concludes the paper with recommendations for future work.

## 2 Background and Related Work

In recent years, there has been growing interest in the use of computer vision techniques for construction site monitoring. Early work explored the use of computer vision in the AEC industry [12], focusing on the benefits and challenges of implementing these technologies. By incorporating object detection, tracking, and activity recognition in real-time site monitoring computer vision techniques gained popularity [13].

Subsequent research investigated the automatic detection of construction site safety hazards using computer vision [14], further emphasizing the potential of these methods in improving safety and efficiency on construction sites. In parallel, a review of computer vision methods for 3D construction site modelling was conducted [15], highlighting the usefulness of techniques such as Structure-from-Motion (SfM) and photogrammetry for generating 3D models from images or video data collected on-site.

More recent projects have critically reviewed automated visual recognition of construction site objects [16] and investigated smart construction site monitoring systems based on Unmanned Aerial Vehicles (UAVs) and artificial intelligence [17]. These studies demonstrated the effectiveness of combining AI technologies, including computer vision algorithms, with UAVs for progress monitoring, safety analysis, and resource management. Latest research includes a comprehensive review of object detection and tracking in the construction industry [18], identifying the challenges that need to be addressed to improve the effectiveness of these techniques in construction site monitoring.

### 2.1 The Use Case: Progress Monitoring for Window Installations

The above projects provide valuable insights and methodologies, which are applicable and adaptable for our current project, which aims to develop CPM for the window installation process based on object detection modelling.

The installation of windows can be monitored from the outside of buildings, e.g. by using drones and from the inside, i.e. from the room to which the window belongs to.

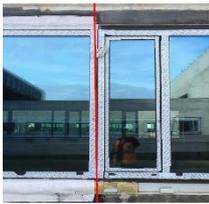 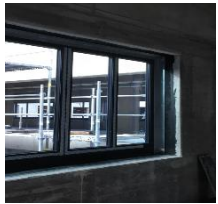 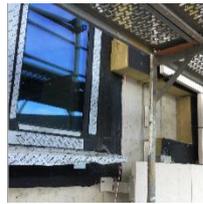

**Fig. 1**: Windows in rough construction        **Fig. 2**: Windows in façade ("peel away")

**Fig. *1*** (left) depicts a sealed (left) and an unsealed (right) window from the outside. The right picture of **Fig. *1*** represents an installed window from the inside. **Fig. *2*** (left) presents an installed window with external insulation and weather protection layer under installation plus the scaffold in front. **Fig. *2*** (right) represents a window from the inside with acoustic insulation layer added to the load-bearing wall.

## 3        Object Detection Algorithm

Object detection is a pivotal computer vision task that involves identifying and localizing instances of objects within images or videos. Applications exist in various fields, such as autonomous vehicles, robotics, and construction progress monitoring [19], [20], [21]. Recently, convolutional neural networks (CNNs) have significantly improved object detection capabilities by automatically learning hierarchical features from raw pixel data, offering higher accuracy and better generalization [22], [23].

The latest version of the YOLO object detection algorithms, renowned for their real-time processing capabilities and competitive accuracy, incorporates several architectural improvements and optimization techniques that enhance both accuracy and speed [24], including:

- a modified backbone architecture based on CSPDarknet53 with Cross Stage Hierarchical (CSH) connections,
- an enhanced feature pyramid network for better handling of objects with varying sizes and aspect ratios,
- optimized anchor boxes tailored to the specific object classes and aspect ratios present in the training dataset, and
- Mosaic data augmentation technique that exposes the model to a diverse set of object scales, orientations, and lighting conditions. Additionally, it employs an improved loss function for more accurate and stable training and boasts an efficient implementation using CUDA and cuDNN libraries.

### 3.1        Object-Identification for Window Installation

In the context of CPM and quality management, the documentation of window installations plays a crucial role in allowing stakeholders to visually assess ongoing work [25]. Timely and accurate CPM for windows is essential for maintaining construction activities' progress according to schedule and quality requirements [26], including the avoidance of so-called 'hidden' errors and omissions.

Improved accuracy in object detection provides advanced capabilities for the identification of missing parts, the determination of the correct window type, and the verification of assembly tolerances. The use of object-detection algorithms can enhance safety by replacing manual monitoring processes with automated visual documentation using either integrated devices or UAV. The risk of accidents and injuries associated with quality management tasks on complex construction sites is reduced as well.

The outdoor object detection for windows uses a rather conservative approach, since polished or glassed surfaces pose several problems to objection detection algorithms. Window frames often have only a limited number features to be detect and classified. Finally, different visual effects of the materials used in the installation process (e.g. color of sealing tapes) make a fully automatic detection process error prone. Therefore, it was decided to mark each window with a unique QR-code sticker. Provided a sufficient image quality, QR-codes can be detected and located robustly. For outdoor detection of windows installed, we envisage to use an active learning approach.

**Table 1.** Comparison of Indoor versus Outdoor Features for CPM of window installation

|  | Feature | Indoor | Outdoor |
|---|---|---|---|
| **General Features** | Parties involved | 4 | 6 |
|  | Access to Quality Mgr. | yes | scaffold required |
| **Object-Identification Feature** | Parts for identification | handles, hinges, actuators, etc. | nearly none |
|  | Obstructions | limited | scaffold, powerlines |
|  | Objects in background | depends on site | multiple |
|  | Mirroring effect | limited | high, due to reflective surfaces |

In summary, the application of advanced object detection algorithms for CPM, particularly for the indoor part of CPM for windows, provides several benefits, such as:

- automatic detection and localization of windows' position using real-time processing capabilities,
- to receive up-to-date information about the status of all steps of the window installation process (see Fig.3),
- to facilitate proactive decision-making in case of detected anomalies,
- fast identification of the actors responsible for corrections and
- instantaneous information provision for actors through electronic media.

Thus, automated, AI-based CPM contributes to the high-quality completion of construction projects by facilitating early detection of deviations, timely decision-making for necessary corrections and ensuring compliance with project schedules.

## 4 Dataset Creation and Annotation

### 4.1 Progress Monitoring and Checkpoints Identification

Data collection is a crucial component of the project, as it lays the foundation for subsequent model training and prediction. A comprehensive dataset consisting of high-quality images is essential for achieving accurate and reliable results in object detection.

**In case of indoor object identification**: We compiled a representative sample of images covering the range of window construction scenarios to be monitored in the future. We conducted the data collection in January 2023 at the Beyer-Bau between 11 a.m. and 12 p.m. The dataset comprises 347 images, stored in JPG format with a resolution of 3060x4080 pixels. We considered various factors during image capture, including window types, camera position, lighting conditions, and foreground and background diversity. The project's success will depend on dataset's quality and relevance.

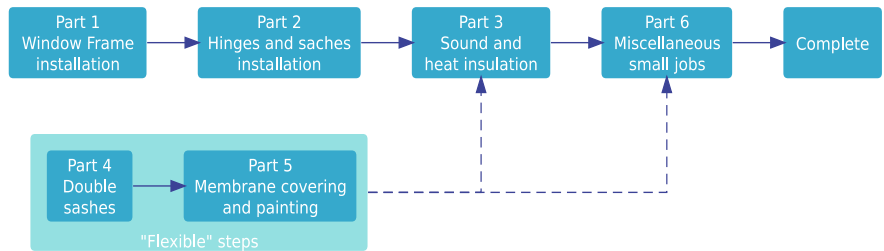

**Fig. 3**. Window installation process

After analysing project requirements, it's clear that process monitoring involves tracking completion percentage and operation order. Fig. 3 describes the window installation process and divides the installation into six parts with corresponding completion percentages. However, an additional subprocess is needed to address the "flexible" steps, such as the combined work of Part 4 and Part 5. Workers must decide whether to perform these tasks before or after completing Part 3, depending on material availability and construction system coordination. The window installing percentage, as shown in Table. 1, can be divided into eight "check points," determined by the completion status of the corresponding steps.

By combining data collection, monitoring plan determination, and data preprocessing, we created an efficient and comprehensive framework to monitor window construction progress using YOLO. This approach simplifies the task and ensures the proper characterization of the window installation process while preserving the necessary details for object detection.

**In case of outdoor object identification**: The task is to compare a list of target-states of building parts with their as-is-states. In our case this is a list of windows together with their planned position and orientation, which is derived automatically from BIM-data. Each window is assigned to a unique ID which is printed on a sticker and attached to the corresponding window. The ID is also given in the list. The positions and orientations are given relative to a coordinate system attached to the building, which is marked on the carcass by additional QR-code-signs, the positions of the coordinate system markers are given in the list as well.

The second input data set is a large number of images taken by a drone. The images must show a considerable overlap of the depicted scene that would allow rendering a 3D point cloud.

### 4.1    Annotation and Label Formatting

Using makesense.ai, a web-based platform for creating labelled datasets, we manually annotated the dataset. The platform provides an online tool for annotating images, as well as powerful tools for organizing labelled datasets and collaborating on large-scale projects. We selected makesense.ai for its flexibility, ease of use, and advanced sharing capabilities.

The labels were exported to YOLO format with one TXT file corresponding to each image. These files contain information on the class of the bounding box, and the x and y coordinates of the centre point, as well as the width and height of the bounding box in normalized xywh format. Class numbers are zero-indexed, and box coordinates are in normalized format with values ranging from 0 to 1.

**Table 2.** Window installation percent

| Completion Percentage | Description |
|---|---|
| 20% | Part 1: Secure pre-assembled window frame into the wall opening, use support materials to prevent frame creeping and ensure stability. |
| 40% | Part 1 + Part 2: Attach hinges to the frame, install sashes, and test tightness for smooth operation and no leaks. |
| 60% | Part 1 + Part 2 + Part 4: Install inner sashes for double-layered windows or count single-layered windows as 40% complete. |
| 65% | Part 1 + Part 2 + Part 3: Apply adhesive to fill gaps from Part 1, complete waterproofing, soundproofing, and heat insulation. Seal gaps with plastic strips. |
| 70% | Part 1 + Part 2 + Part 3: Apply adhesive to fill gaps from Part 1, complete waterproofing, soundproofing, and heat insulation. Seal gaps with plastic strips. |
| 85% | Part 1 + Part 2 + Part 3 + Part 4: Complete gap filling, waterproofing, soundproofing, and heat insulation; install inner sashes for double-layered windows. |
| 95% | Part 1 + Part 2 + Part 3 + Part 4 + Part 5: Complete painting and install inner sashes for double-layered windows. |
| 100% | Part 1 + Part 2 + Part 3 + Part 4 + Part 5 + Part 6: Remove plastic membrane from glass, complete any final miscellaneous tasks. |

### 4.2    Data Augmentation and Dataset Partitioning

We applied various image-level augmentations to enhance the dataset's diversity, including horizontal flip, rotation (clockwise and counterclockwise 90° and ±15°), shearing (±15°), brightness adjustment (±25%), and blurring (2px). Consequently, the final dataset contained 768 images.

To create appropriate subsets for training, validation, and testing, we partitioned the dataset into three sets: training (729 images, 88%), validation (52 images, 6%), and testing (52 images, 6%).

In summary, this chapter described the dataset creation and annotation process for monitoring window construction progress using YOLO. Through careful image selection, defining monitoring plans, and pre-processing data, we successfully developed a robust dataset for training and testing object detection models. The integration of the "Percentage" and "Process" subsystems allowed for a more accurate representation of the window installation process, while proper annotation and data augmentation ensured the quality and diversity of the dataset. This comprehensive approach should provide a solid foundation for the effective application of the YOLO object detection algorithm in monitoring window construction progress, leading to accurate and reliable results in real-time performance.

### 4.3 Implementation

**Indoor CPM of window installation**: The object detection model was implemented using the YOLO algorithm. To optimize the model's performance, various hyperparameters were carefully selected. The chosen epoch count was 1000, the image size was set to 640, and the batch size was 32. The Adam optimizer was utilized with an initial learning rate of 5e-4 and a final learning rate of 1e-3. Plotting was enabled to visualize the training progress.

For the training process, a powerful hardware setup was employed, which included an NVIDIA A100-SXM4-40GB GPU, a combination of 11 Intel(R) Xeon(R) CPU with 2.20GHz, and 85.3GB of RAM. This setup ensured a smooth and efficient training experience, allowing the model to learn from the provided dataset effectively.

The dataset used for training, collected in January 2023 at Beyer-Bau, was preprocessed and annotated to create a comprehensive training set. The dataset comprised 347 high-resolution images, capturing various window construction scenarios, camera positions, lighting conditions, and foreground and background diversity. The training set enabled the model to learn and differentiate between the various construction stages effectively.

During the implementation, the YOLO-based object detection model was designed to recognize critical checkpoints of the construction progress, such as the completion percentages and flexible steps in the process. This innovative approach allowed the model to monitor the window installation process effectively and efficiently.

The implementation also involved the creation of a monitoring plan, which combined the "Percentage" and "Process" subsystems aiming to effectively characterize the window installation process. By transforming the continuous and streamlined work of window installation monitoring into the detection of windows under different situations of time sections during the construction process, the entire task was simplified into a condition more suitable for the YOLO algorithm.

**Outdoor CPM of window installation**: The algorithm for object identification comprises of three steps, including:

1. In the first step all images are examined with respect to suitability, i.e. sharpness, contrast and exposure. Then the images are registered, which means that the geographic coordinate system is established, For each image the camera position as well as the position of key features are computed.

2. QR-code-detection and localization is carried out for each image. Recognized QR-codes are checked against the list of target-states of each window. If a matching window-ID is found, the position is checked and the detected window-ID is removed from the list.

3. The result of step 2 is a list of windows, which could not be detected by their QR-codes. There are many possible reasons for this: The necessary images might not be taken or delivered, the technical quality of some images might be poor, the window might be hidden by some other structure, the QR-code-sticker might have fallen of etc.. Some of the reasons can be checked automatically. Regardless of the result, further human activities must be taken to either verify the state of the windows with alternative means or to request further image data, necessary to finalize the outdoor object detection.

## 5 Results and Discussion

### 5.1 Results

After completing the training process, the model achieved its best performance at epoch 82, with the best model saved as best.pt. An early stopping mechanism was employed due to no improvement observed in the last 50 epochs, which allowed for efficient use of computational resources. In total, 132 epochs were completed in 0.227 hours.

The evaluation of the trained model was carried out on a separate validation dataset. The model demonstrated promising results (see Fig. 4) with an overall mAP50 (mean Average Precision at 50% IoU) of 0.953 and an mAP50-95 (mean Average Precision at 50% to 95% IoU) of 0.678. These metrics indicate a high degree of accuracy in detecting windows at different stages of the construction process. The model also showed high precision and recall scores (see Fig. 5) across the various completion percentages and flexible steps, suggesting a good understanding of the different stages and their respective checkpoints.

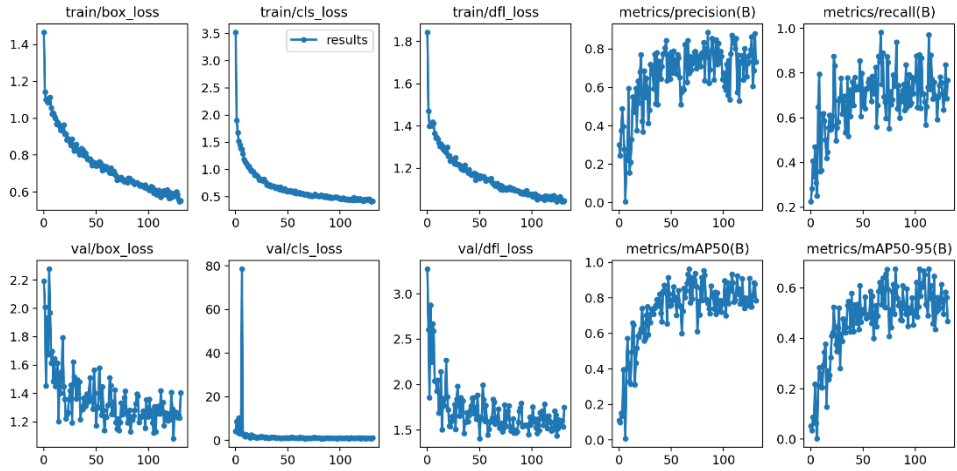

**Fig. 4**. Training Results

### 5.1 Discussion

The results obtained from the trained YOLO-based object detection model indicate its potential for monitoring window construction progress effectively. The model successfully identified various checkpoints and completion percentages in the window installation process, providing accurate and timely information on the construction progress. The high mAP scores across different IoU thresholds demonstrate the model's robustness and ability to generalize well to new data.

The use of the YOLO algorithm for this task proved to be a suitable choice, as it allowed for a streamlined and efficient approach to process monitoring. The combination of the "Percentage" and "Process" subsystems provided a comprehensive characterization of the window installation process, enabling the model to capture the necessary details for accurate object detection.

One potential limitation of the current implementation is the reliance on a single dataset for training, which may not cover all possible scenarios and variations in window construction. To improve the model's generalizability further, it would be beneficial to include more diverse data from various construction sites and time periods.

In conclusion, the YOLO-based object detection model for monitoring window construction progress has demonstrated promising results, indicating its potential for practical application in the construction industry. Future work could involve testing the model on real-world construction sites to evaluate its performance in real-time and exploring the integration of the model with other monitoring systems to provide a comprehensive solution for construction progress tracking.

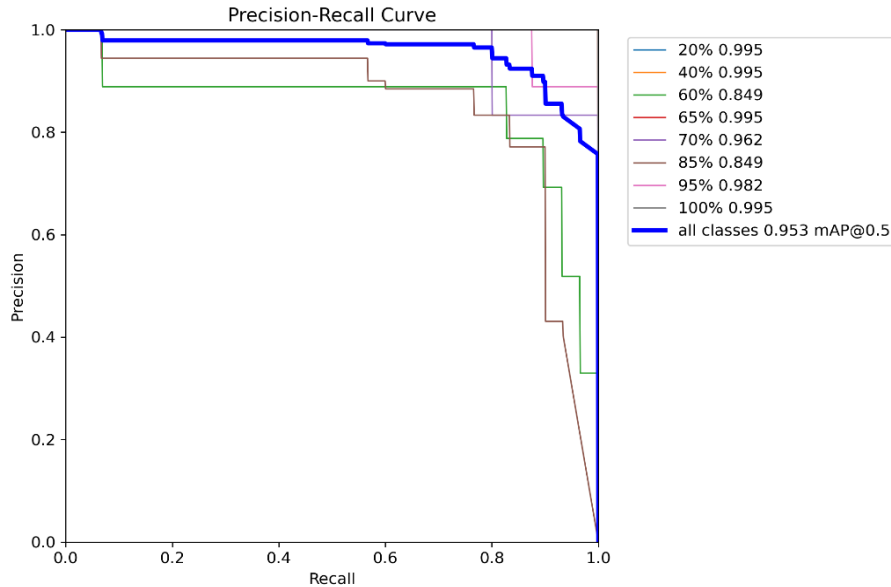

**Fig. 5**. Precision and recall score

# 6      Conclusion and Future Work

## 6.1      Conclusion

This study aimed to develop an efficient and effective method for CPM of window installations. The project involved data collection, pre-processing, model training, evaluation, and analysis. The results demonstrated that the trained model for indoor object detection could accurately identify various checkpoints and completion percentages in the window installation process.

The YOLO algorithm's adoption for this task proved to be a suitable choice, as it allowed for a streamlined and efficient approach to process monitoring. The combination of the "Percentage" and "Process" subsystems provided a comprehensive characterization of the window installation process (see **Fig. *6***), enabling the model to capture the necessary details for accurate object detection.

Despite some limitations in the current implementation, such as the reliance on a single dataset for training, the YOLO-based object detection model for monitoring window construction progress has shown promising results, indicating its potential for practical application in the construction industry.

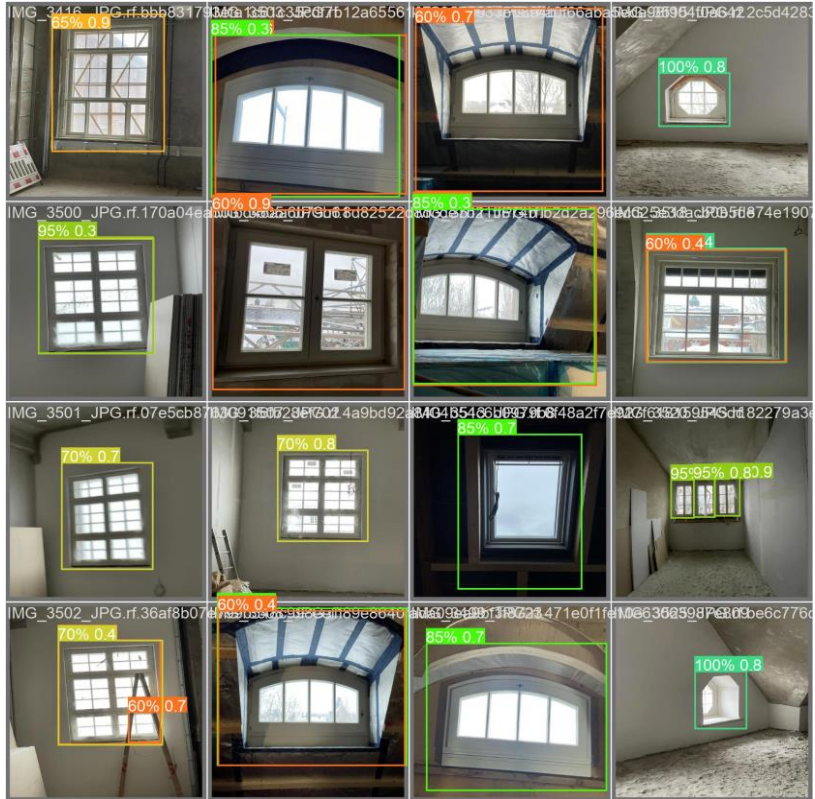

**Fig. 6.** Implementation of trained model

## 6.2 Future Work

To further enhance the model's performance and applicability, several avenues for future work could be explored:

Expand the training dataset: Collecting data that are more diverse would improve the model's ability to adapt to different scenarios and variations in window construction.

Real-world testing: Evaluating the model's performance on real-world scenarios will provide insights into its effectiveness and help identify areas for improvement.

Integration with other monitoring systems: Exploring the integration of the YOLO-based model with other monitoring systems, such as site management software or Building Information Modelling, could provide a comprehensive solution for CPM.

Adaptation to other construction tasks: Investigating the applicability of the YOLO-based object detection model for monitoring other construction tasks, such as masonry, or steel erection could broaden the model's use in the construction industry.

By pursuing these future work directions, the YOLO-based object detection model proposed in this paper could become a valuable tool in the construction industry.

# 7 Acknowledgements

We would like to express our appreciation to Mr. Fang Jian for his contribution to the project. His knowledge and dedication have been instrumental in advancing our understanding of the subject matter and achieving our research objectives.

The publication is part of the research project entitled "iECO – Intelligence Empowerment of Construction Industry" which receives funding from Bundesministerium für Wirtschaft und Klimaschutz (BMWK) based on a resolution of the German Bundestag. Authors gratefully acknowledge the support and funding from the BMWK. The content of this publication reflects the author view only and the BMWK is not responsible for any use that may be made of the information it contains.

# 8 References

1. Allan L, Menzel K (2009) Virtual Enterprises for Integrated Energy Service Provision. In: Camarinha-Matos LM, Afsarmanesh H, Paraskakis I (eds) Leveraging Knowledge for Innovation in Collaborative Networks: 10th IFIP WG 5.5 Working Conference on Virtual Enterprises, PRO-VE 2009, Thessaloniki, Greece, October 7-9, 2009. Proceedings, vol 307. Springer-Verlag Berlin Heidelberg, Berlin, Heidelberg, pp 659–666

2. Ahmed A, Ploennigs J, Gao Y et al. (2009) Analysing building performance data for energy-efficient building operation. In: Dikbas A, Ergen E, Giritli H (eds) Managing IT in Construction/Managing Construction for Tomorrow. Chapman and Hall/CRC, Boca Raton, pp 211–220

3. Manzoor F, Linton D, Loughlin M et al. (2012) RFID based efficient lighting control. International Journal of RF Technologies 4:1–21. https://doi.org/10.3233/RFT-2012-0036

4. Menzel K, Tobin E, Brown KN et al. (2009) Performance Based Maintenance Scheduling for Building Service Components. In: Camarinha-Matos LM, Afsarmanesh H, Paraskakis I (eds) Leveraging Knowledge for Innovation in Collaborative Networks: 10th IFIP WG 5.5 Working Conference on Virtual Enterprises, PRO-VE 2009, Thessaloniki, Greece, October 7-9, 2009. Proceedings, vol 307. Springer-Verlag Berlin Heidelberg, Berlin, Heidelberg, pp 487–494

5. Menzel K, Eisenblätter K, Keller M et al. (2002) Context-sensitive process and data management on mobile devices. In: Turk Z, Scherer RJ (eds) eWork and eBusiness in Architecture, Engineering and Construction: Proceedings of the 4th European Conference, Portoroz, Slovenia. Swets & Zeitlinger Publishers, Lisse, The Netherlands, pp 549–554

6. Martin Keller, Karsten Menzel, Sven-Eric Schapke et al. (2007) Framework zur Referenzmodellierung im Bauwesen. In: Loos P (ed) Kollaboratives Prozessmanagement: Unterstützung kooperations- und koordinationsintensiver Geschäftsprozesse am Beispiel des Bauwesens, 1st edn. Logos-Verl., Berlin, pp 105–124

7.  Mohan N, Gross R, Menzel K et al. (2021) Opportunities and Challanges in the Implementation of Building Information Modelling for Prefabrication of Heating, Ventilation, and Air Conditioning Systems in Small and Medium Sized Contracting Companies in Germany: a Case Study. In: Casares J, Mahdjoubi L, Garrigos AG (eds) WIT Transactions on the Built Environment: BIM 2021. WIT Press, [S.l.], pp 117–126

8.  Karlapudi J, Menzel K, Törmä S et al. (2020) Enhancement of BIM Data Representation in Product-Process Modelling for Building Renovation. In: Nyffenegger F, Ríos J, Rivest L (eds) Product Lifecycle Management Enabling Smart X: 17th IFIP WG 5.1 International Conference, PLM 2020, Rapperswil, Switzerland, July 5–8, 2020, Revised Selected Papers, 1st ed. 2020, vol 594, pp 738–752

9.  Valluru P, Karlapudi J, Mätäsniemi T et al. (2021) A Modular Ontology Framework for Building Renovation Domain. In: Camarinha-Matos LM, Boucher X, Afsarmanesh H (eds) Smart and Sustainable Collaborative Networks 4.0, vol 629. Springer International Publishing, Cham, pp 323–334

10. Valluru P, Karlapudi J, Menzel K et al. (2020) A Semantic Data Model to Represent Building Material Data in AEC Collaborative Workflows. In: Camarinha-Matos LM, Afsarmanesh H, Ortiz A (eds) Boosting Collaborative Networks 4.0: 21st ifip wg 5.5 working. Springer International Publishing, Cham

11. Menzel K, Törmä S, Markku K et al. (2022) Linked Data and Ontologies for Semantic Interoperability. In: Daniotti B, Lupica Spagnolo S, Pavan A et al. (eds) Innovative Tools and Methods Using BIM for an Efficient Renovation in Buildings. Springer International Publishing, Cham, pp 17–28

12. Rashid KM, Louis J (2019) Times-series data augmentation and deep learning for construction equipment activity recognition. 1474-0346 42:100944. https://doi.org/10.1016/j.aei.2019.100944

13. Shen R, Huang A, Li B et al. (2019) Construction of a drought monitoring model using deep learning based on multi-source remote sensing data. 1569-8432 79:48–57. https://doi.org/10.1016/j.jag.2019.03.006

14. Zhang Y, Yuen K-V (2022) Applications of Deep Learning in Intelligent Construction. In: Cury A, Ribeiro D, Ubertini F et al. (eds) Structural Health Monitoring Based on Data Science Techniques. Springer International Publishing, Cham, pp 227–245

15. Liu J, Luo H, Liu H (2022) Deep learning-based data analytics for safety in construction. Automation in Construction 140:104302. https://doi.org/10.1016/j.autcon.2022.104302

16. Mahami H, Ghassemi N, Darbandy MT et al. Material Recognition for Automated Progress Monitoring using Deep Learning Methods. Accessed 21 Apr 2023

17. Xiong W, Xu X, Chen L et al. (2022) Sound-Based Construction Activity Monitoring with Deep Learning. 2075-5309 12:1947. https://doi.org/10.3390/buildings12111947

18. Elghaish F, Matarneh ST, Alhusban M (2021) The application of "deep learning" in construction site management: scientometric, thematic and critical analysis. 1471-4175 22:580–603. https://doi.org/10.1108/CI-10-2021-0195

19. Deng J, Dong W, Socher R et al. (2009) ImageNet: A large-scale hierarchical image database. In: 2009 IEEE Conference on Computer Vision and Pattern Recognition, pp 248–255

20. Rich Feature Hierarchies for Accurate Object Detection and Semantic Segmentation IEEE Conference Publication IEEE Xplore. https://ieeexplore.ieee.org/document/6909475. Accessed 21 Apr 2023

21. Redmon J, Divvala S, Girshick R et al. (2016) You Only Look Once: Unified, Real-Time Object Detection. In: 2016 IEEE Conference on Computer Vision and Pattern Recognition (CVPR), pp 779–788

22. He K, Gkioxari G, Dollár P et al. (2017) Mask R-CNN. In: 2017 IEEE International Conference on Computer Vision (ICCV), pp 2980–2988

23. Ren S, He K, Girshick R et al. Faster R-CNN: Towards Real-Time Object Detection with Region Proposal Networks. In: Advances in Neural Information Processing Systems, vol 28. Curran Associates, Inc

24. Bochkovskiy A, Wang C-Y, Liao H-YM (2020) YOLOv4: Optimal Speed and Accuracy of Object Detection. Accessed 21 Apr 2023

25. Kopsida M, Ioannis B, Vela P (2015) A Review of Automated Construction Progress Monitoring and Inspection Methods

26. Golparvar-Fard M, Peña-Mora F, Savarese S (2011) Integrated Sequential As-Built and As-Planned Representation with D4AR Tools in Support of Decision-Making Tasks in the AEC/FM Industry. Journal of Construction Engineering and Management 137:1099–1116. https://doi.org/10.1061/(ASCE)CO.1943-7862.0000371